\documentclass{bmvc2k}

\usepackage{times}
\usepackage{epsfig}
\usepackage{graphicx}
\usepackage{amsmath}
\usepackage{amssymb}

\usepackage{subfigure}
\usepackage{multirow}
\usepackage{booktabs} 
\usepackage[lined,boxed, ruled]{algorithm2e}
\usepackage{xcolor}
\usepackage{bbding}

\title{MOS: A Low Latency and Lightweight Framework for Face Detection, Landmark Localization, and Head Pose Estimation}

\addauthor{Yepeng Liu}{ericyepeng123@gmail.com}{1}
\addauthor{Zaiwang Gu}{guzaiwang01@gmail.com}{1}
\addauthor{Shenghua Gao}{gaoshh@shanghaitech.edu.cn}{2}
\addauthor{Dong Wang}{dong.wang@ubtrobot.com}{1}
\addauthor{Yusheng Zeng}{bill.zeng@ubtrobot.com}{1}
\addauthor{Jun Cheng\textsuperscript{\dag}}{sam.j.cheng@gmail.com}{31}

\addinstitution{
 UBTech Research,\\
 UBTech Robotics Corp Ltd, China
}
\addinstitution{
 Shanghaitech University,\\
 China
}
\addinstitution{
	Institute for Infocomm Research, \\
	Agency for Science, Technology \\
	and Research, Singapore\\
	\textsuperscript{\dag} Corresponding Author
}

\runninghead{LIU ET AL.}{MOS: A Low Latency and Lightweight Framework}

\begin{document}

\maketitle

\begin{abstract}
With the emergence of service robots and surveillance cameras, dynamic face recognition (DFR) in wild has received much attention in recent years.  Face detection and head pose estimation are two important steps for DFR. Very often, the   pose is estimated after the face detection. However, such sequential computations  lead to higher latency.  In this paper, we propose a low latency and lightweight network  for simultaneous face detection, landmark localization and head pose estimation. Inspired by the observation that it is more challenging to locate the facial landmarks for faces with large angles, a pose loss is proposed to constrain the learning. Moreover, we also propose an uncertainty multi-task loss to learn the weights of   individual tasks automatically.  Another challenge is that robots often use low computational units like ARM based computing core and we often need to use lightweight networks instead of the heavy ones, which lead to performance drop especially for small and hard faces. In this paper, we  propose   online  feedback sampling to augment the training samples across different scales, which increases the diversity of training data automatically.   Through validation in commonly used WIDER FACE, AFLW and AFLW2000 datasets, the results show that   the proposed method achieves the state-of-the-art performance   in low computational resources.  

\end{abstract}

\section{Introduction}
\label{sec:intro}
With the popularization of service robots and surveillance cameras, dynamic face recognition (DFR) in the wild has become widely used.  DFR is different from static face recognition (SFR). In SFR  such as Apple face ID, the aim is to recognize faces within a specific area or range with relatively high requirements for the angle, distance, and position, which often requires human cooperation. However, in DFR applications such as surveillance, the aim is  to automatically recognize persons walking in natural forms, generally requiring no human cooperation.   The traditional pipeline for DFR is to: 1) face detection and tracking; 2) head pose estimation and face image quality evaluation; 3)  image selection based on the head pose and  quality; 4) face identity comparison.  The face recognition  can be deployed on  device  with low computing power  or cloud side. However,  cloud service is often interrupted by unstable network connection and limited bandwidth.   Therefore, real-time face detection and head pose estimation in mobile side  become  necessary for DFR in the wild.

Face detection is a prerequisite step of facial image analysis such as facial attribute, e.g., expression ~\cite{zhang2018joint}, age ~\cite{pan2018mean} and  face identity  ~\cite{schroff2015facenet, liu2017sphereface, deng2019arcface}. With the recent development of deep learning, significant improvements have been achieved in face detection by utilizing   CNN-based object detectors. 
Among these methods,   single-stage based deep learning methods have shown to be promising ~\cite{zhang2018single,zhang2017s3fd,najibi2017ssh,tang2018pyramidbox}. The single-stage methods detect  the faces of different scales without significant increment of time-consumption even when there is an increased number of faces in the images. These methods densely sample face locations and scales on feature pyramids ~\cite{liu2016ssd, lin2017feature}, demonstrating promising performance and yielding high speed. Very often, the five facial landmarks are localized simultaneously ~\cite{deng2019retinaface}. However, landmark localization is still insufficient for pose estimation in DFR due to poor accuracy for face with large angles. In this work, we   propose a novel low latency and lightweight  framework for real-time face detection, landmark localization and   pose estimation. To avoid a large increment in computation, we propose to detect face, locate the landmarks and estimate the head pose simultaneously. 

There are three major challenges. 
Firstly, it is not an easy task to learn face detection, facial landmark localization and head pose estimation simultaneously and get accurate results for all tasks with low computational resources ~\cite{ranjan2017hyperface, deng2019retinaface}. 
Secondly, the existing datasets for face detection, facial landmark localization and head pose estimation are  non-unified and inconsistent of the scales for joint training while it is difficult to obtain accurate pose annotations, especially for small faces ~\cite{zhuang2019fldet}. 
Lastly, false positive in face detection affects the user experience and it is very challenging to reduce it with low computational resources.

Intuitively, attention area by pose estimation shall be consistent with that by face classification.  Therefore, we train a multi-task network and utilize the feature maps to validate our assumption visually. The result shows that the attention area for pose estimation  and classification  are highly overlapped. This motivates us to  integrate the pose estimation into face detection such that it may maintain or even improve  the  accuracy while reducing extra step for pose estimation. 
In addition, we observe that the landmark localization is more difficult in faces with large angles and propose an additional constraint  called pose loss to regularize the training.  
In order to train a model for simultaneous face detection and head pose estimation, we further label the pose of each face for the WIDER FACE dataset, which will be released. 


The main contributions  are summarized as follows:
\begin{enumerate}
	
	\item We propose a low latency and lightweight architecture for face detection, facial landmark localization and head pose estimation simultaneously.   
	\item  
    We propose uncertainty multi-task loss to make the face detection, landmark localization  and head pose estimation more accurate. 
	\item  
	We propose an online data-feedback  augmentation strategy, which  improves the data balance in face detection.
	\item  
	The   results show that our method outperforms  other lightweight methods, especially for the \textit{hard} subset.
	
\end{enumerate}
\section{Related Work} \label{related}
\begin{figure*}
	
	\centering
	\includegraphics[width=5in]{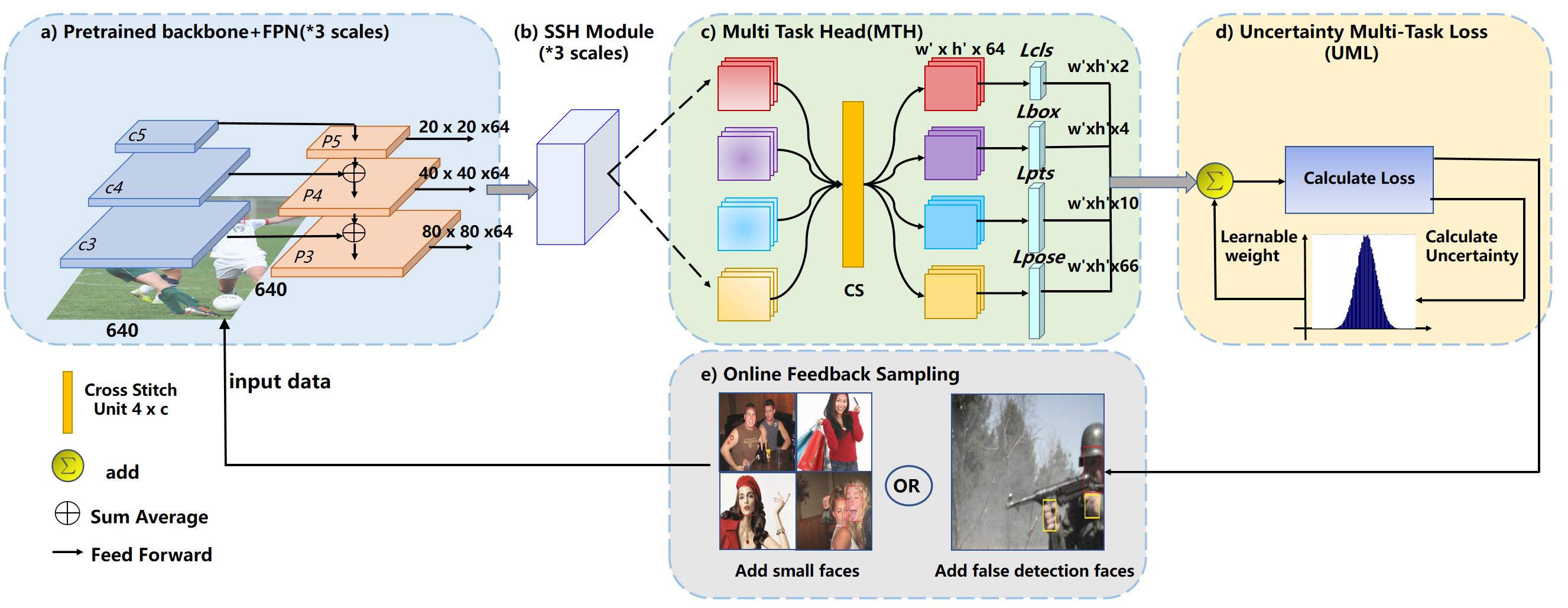}
	\caption{ Overview of the proposed method.  MOS adopts the feature pyramid from $P3$ to $P5$ followed by  SSH context head module and a multi-task head module.   A cross stitch unit ($n\times c$, $n$ denotes task number and $c$ denotes channel number) is used in the multi-task head to combine the feature maps linearly for face detection, landmark localization and head pose estimation. 
	}
	\label{fig2}
\end{figure*}


\textbf{Face detection: }Face detection ~\cite{Yang2016WIDER,jain2010fddb} has been a hot topic in the past decades. Earlier methods ~\cite{viola2004robust} for face detection are mostly based on hand-crafted features. 
Existing methods are divided into two categories: two-stage methods (e.g. Faster R-CNN ~\cite{ren2015faster}) and single-stage methods (e.g. SSD ~\cite{liu2016ssd}). 
Single-stage methods have great advantages in inference speed, though these detectors often lead to  higher false positive. 
There are some methods ~\cite{liu2020hambox,chi2019selective} that reduce the false positive detection by improving the quality of the matching box. Another problem is how to detect small faces. S3FD ~\cite{zhang2017s3fd} used multiple strategies to improve the performance of small faces. ProgressFace ~\cite{zhuprogressface} proposed a novel scale-aware progressive training mechanism to address large scale variations across faces. 

\textbf{Head pose estimation: }Head pose estimation   has been widely studied.   Early method uses the facial landmarks ~\cite{kazemi2014one, bulat2017far, zhu2016face, kumar2017kepler} to estimate the head pose. 
The methods based on facial landmarks and Perspective-n-Point (PnP) ~\cite{pnp1981} are very popular because there is no need to include a pose estimation model. However, the error in landmark localization will be propagated. Recently, CNN based methods for direct pose estimation made some progress. 
Hopenet ~\cite{2018Fine} combined ResNet50 with a multi-loss function, which is composed of  a classification loss and regression loss. 
FSAnet ~\cite{2019FSA} proposed to learn a fine grained structure mapping for spatially grouping features before aggregation. Although some of these methods ~\cite{hsu2018quatnet,zhou2020whenet,dai2020rankpose} achieve excellent results on pose estimation, none of them is combined with face detection. 

\textbf{Multi-task learning: }
The recent progress  in multi-task learning  mainly focuses  on the shared network architecture ~\cite{misra2016cross} and the loss function weights of the tasks  ~\cite{kendall2018multi}. In MTAN ~\cite{MTAN2019}, Liu \textit{et al.} proposed  a new network based on SegNet ~\cite{badrinarayanan2017segnet}  and obtained the state-of-the-art performance on the tasks of semantic segmentation and depth estimation on the outdoor CityScapes dataset ~\cite{cordts2016cityscapes}. 
There are some methods attempting to solve the problem of face detection and alignment in one model. In MTCNN ~\cite{zhang2016joint}, Zhang \textit{et al.}  used a cascaded architecture with three stages of shallow networks to predict face and landmark locations in a coarse-to-fine manner. In RetinaFace ~\cite{deng2019lightweight}, Deng \textit{et al.}  manually annotated five facial landmarks on the WIDER FACE dataset and observed significant improvement in \textit{hard} subset with the assistance of this extra supervision. There are also some efforts to include head pose estimation and other face attributes~\cite{zhang2014facial,ranjan2017hyperface}. 
Ranjan \textit{et al.} ~\cite{ranjan2017hyperface} presented an algorithm for simultaneous face detection, landmark  localization, head pose estimation and gender recognition, which shows that the  multi-task learning can get better results. However, these methods use  hard parameter connection, which limits the feature sharing among different tasks.


\section{Method}

\label{method}
In this paper, we propose a low latency and lightweight framework including the following sub-tasks: face classification, bounding box regression, landmark regression and head pose estimation. Fig. \ref{fig2} shows the overview of the proposed method.

\subsection{Multi-Task Head}

\begin{figure}[htb]
	\centering
	{\subfigure[The different connection method of head modules]{ 
			\includegraphics[width=2.1in,height=2in]{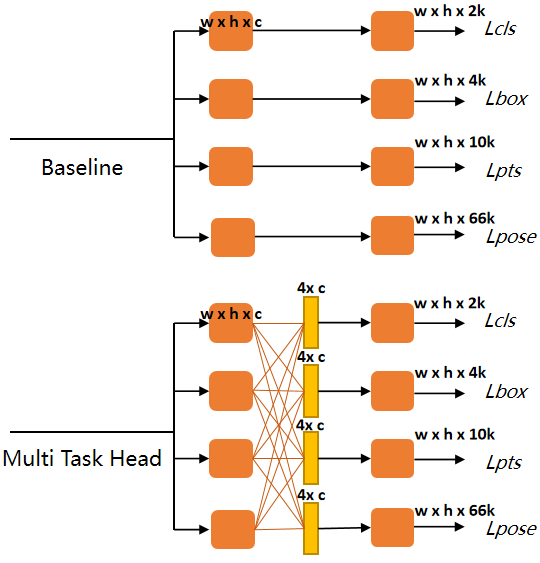} \label{fig3a}
	}}  
	{\subfigure[ The pipeline of online feedback sampling]{ 
			\includegraphics[width=2in,height=2in]{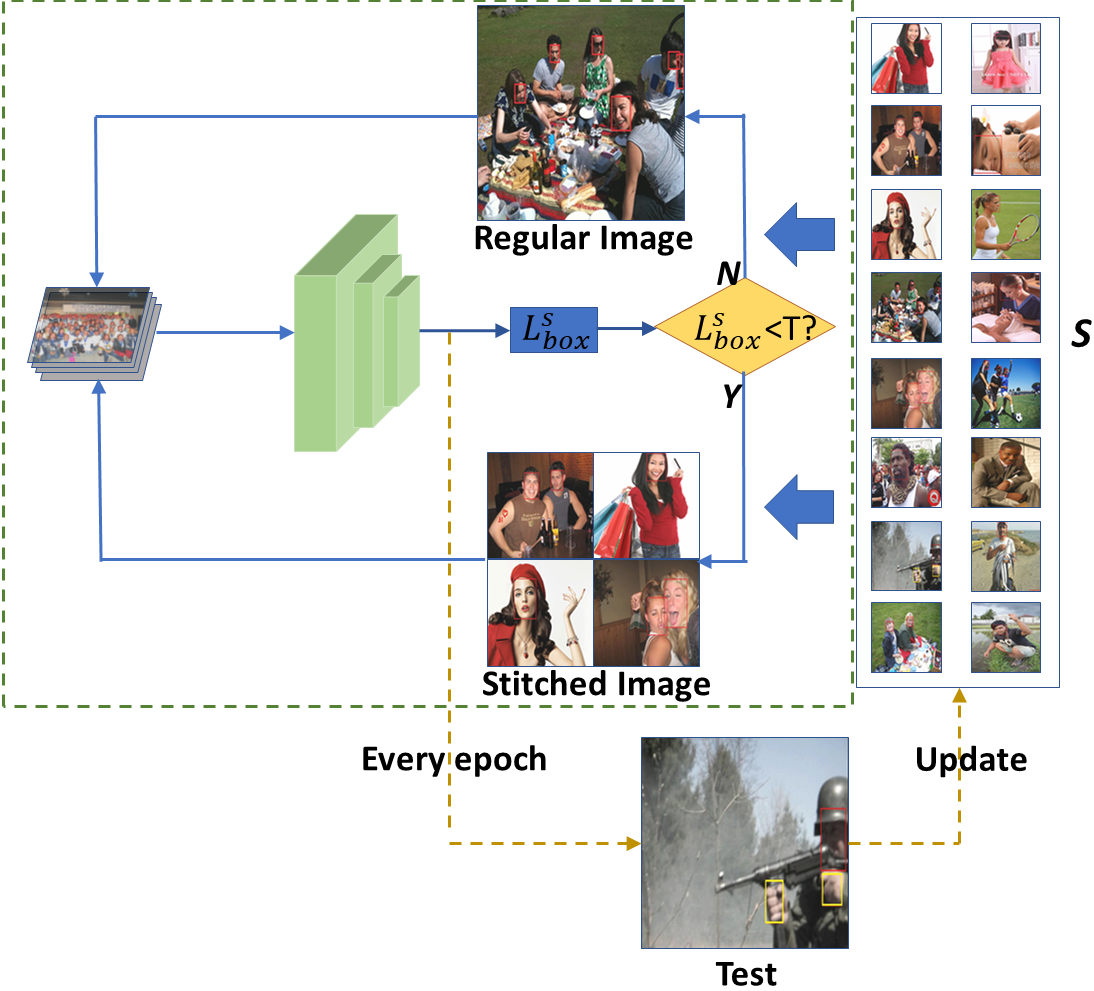} \label{fig3b}
	}} 
	\caption{(a) The top  figure is the baseline head module with hard parameters connection; The bottom is the proposed MTH module with cross-connection. (b) Online Feedback Sampling: Compute $L_{box}^s$ for faces no larger than $25\times 25$ and apply stitching accordingly after every iteration, and value $T$ is set to 0.35; Update the set $S$ after every epoch. The yellow and red boxes in the test image refer to false detection and correct detection respectively. The images will be added into the set $S$ according to the number of false detection boxes.  
	}
	\label{fig3}
\end{figure}

Most of the multi-task branches of face detection \cite{deng2019retinaface,ranjan2017hyperface,zhang2016joint}  are  forked from the last layer  directly, and therefore these sub-tasks actually share all the previous features.
Fig.  \ref{fig3a} shows the baseline head module \cite{deng2019retinaface, ranjan2017hyperface} with hard parameter connection. 
 Although some correlation exists between them, there are still many differences among the sub-tasks. 
For example, landmark regression pays more attention to the location of each landmark while bounding box regression pays more attention to the edge of face area \cite{zhuang2019fldet}.  

Inspired by the above, we propose   a novel head sharing unit called multi-task head (MTH).
 The MTH units try to find the best shared representations for multi-task learning. Different heads   share representations through linear combinations  and learn the optimal  combinations for these tasks.
Fig.  \ref{fig3a} shows the structure of MTH  Different from the baseline head module \cite{deng2019retinaface, ranjan2017hyperface} with hard parameter connection, we further include a cross stitch unit size of $n\times c$  in Fig. \ref{fig3a}, where $n$ denotes the number of tasks and $c$ denotes the number of channels. It computes a linear combination of feature maps followed by additional $1\times 1$ convolutions.   
It shall be noted that the cross stitch unit here is different from that in \cite{misra2016cross}. In our unit, we use channel-wise weights in each channel while shared values were used in \cite{misra2016cross}.

\subsection{Multi-task Loss}

\subsubsection{Loss functions used for the tasks} 
\textbf{The classification loss } is a softmax loss for binary classification denoted as  $L_{cls}(p_{i},p_{i}^{*})$, where ${p}_{i}$ is the predicted probability of anchor ${i}$ being a face and  ${p}_{i}^{*}$ is 1 for the positive anchor and 0 for the negative anchor.

\textbf{The bounding box regression loss } is  a Smooth $\ell_1$ loss \cite{girshick2015fast} denoted as  $L_{box}(t_{i},t_{i}^{*})$, where ${t}_{i}=({t}_{x},{t}_{y},{t}_{w},{t}_{h})$ and ${t}_{i}^{*}=({t}_{x}^{*},{t}_{y}^{*},{t}_{w}^{*},{t}_{h}^{*})$ represent the coordinates of the predicted box and ground-truth box associated with the positive anchors  respectively.    

\textbf{The landmark regression loss } $L_{pts}(l_i, l_{i}^*)$ is also based on Smooth $\ell_1$, where $l_i=\{l_{x_1}, l_{y_1}, \cdots, l_{x_5}, l_{y_5} \}$ and
$l_i^*=\{l_{x_1}^*, l_{y_1}^*, \cdots, l_{x_5}^*, l_{y_5}^* \}$  represent the   predicted landmarks and ground truth with the positive anchors respectively.

%

\textbf{Head Pose estimate loss: } Previously, a regression loss $L_{pose}$ based on cross entropy and mean square error is used for pose estimation \cite{2018Fine}. 
\begin{equation} L_{pose}=H{(a,\hat{a})}+\beta \cdot MSE(a,\hat{a}), \end{equation}
where ${H{(a,\hat{a})}}$ denotes the cross entropy, $MSE$ denotes the mean square error, $a$ and ${\hat{a}}$ denote the annotated bins and the predict bins.
Similar to that in \cite{2018Fine}, we use 66 bins, ${\beta}$ is set to 0.001.  


\subsubsection{Uncertainty Multi-task Loss}

The weights in multi-task learning   are set empirically in RetinaFace \cite{deng2019retinaface} and MTCNN \cite{zhang2016joint}. Recently, some work utilizes the uncertainty to estimate the output in the generic object detection \cite{he2019bounding, kendall2018multi}. In this work, we derive a uncertainty multi-task loss (UML) function by maximizing the overall likelihood with the network.

Let $f^{W}(x)$ be the output of a neural network with input $x$ and weights $W$. The multi-task likelihood is defined as the multiplication of single-task probabilities. In this work, we have one classification task and three regression tasks.
\begin{equation}
p(y|f^{W}(x))=p(y_{1}|f^{W}(x))...p(y_{4}|f^{W}(x)), \label{eq3}
\end{equation} 
where $y=(y_{1},...,y_{4})$ denotes the distribution of the four tasks. 

In maximum likelihood estimation, we compute the log likelihood of the model: 
\begin{equation}
\log(p(y|f^{W}(x)))=\sum_{k=1}^{4}\log(p(y_{k}|f^{W}(x)))
\end{equation} 

Following \cite{kendall2018multi}, the classification likelihood is defined as  Boltzmann distribution
function:
\begin{equation}
p(y_{cla}|f^{W}(x))=\mathtt{Softmax}(\frac{1}{T^2}f^{W}(x)),
\end{equation}  
where the subscript ${cla}$ denotes  the classification task. $T$ is temperature of the system. 

The uncertainty loss of classification task ${L_{cla}^*(W,T)}$ is computed as:
\begin{equation}
L_{cla}^*(W,T) =\frac{1}{T^2}L_{cla}(W)+\log(T),
\end{equation}  
The cross entropy loss without coefficient $T$ can be defined as $L_{cla}(W)$, and it represents the loss function of the classification task without the temperature factor $T$.


The likelihood of a regression task can be defined as a Gaussian with the mean given by the model output $N(f^{W}(x),\sigma^{2})$, where $\sigma$ is the model's observation noise parameter.
The uncertainty loss of the regression task $L_{reg}^*(W,\sigma)$ is computed as :
\begin{equation}
L_{reg}^*(W,\sigma)\propto\frac{1}{2\sigma^2}L_{reg}(W)+\log(\sigma), \label{eq7}
\end{equation} 
Where  $L_{reg}(W)$ denotes the $L_2$ norm of the regression task, including two bounding box regression tasks, a landmark regression task and pose estimation task  in this work. 

Applying   (\ref{eq3})-(\ref{eq7})   on the losses of individual tasks, we compute the overall loss function  $L^{*}$   as:
\begin{equation}
\begin{aligned}
L^{*} =\frac{1}{T_1^2}p_i^*L_{cls}+\frac{1}{2\sigma_1^2}p_i^*{L_{box}}+
\frac{1}{2\sigma_2^2}p_i^*{L_{pts}}+
\frac{1}{2\sigma_3^2}p_i^*{L_{pose}}+\log(T_1\sigma_1\sigma_2\sigma_3)
\end{aligned},
\end{equation}
where $T_1$ is the temperature of the classification task, $\sigma_1, \sigma_2, \sigma_3$ are the observation noise of the bounding box detection, landmark regression and pose estimation tasks. These parameters are obtained through learning.

\subsection{Online Feedback Sampling}
Some methods   improve the accuracy of  small face by changing the distribution of face samples. In PyramidBox++ \cite{li2019pyramidbox++}, Li \textit{et al.}  propose   to resize the images into different scales evenly. These strategies are   beneficial for multi-scale sample training, but it needs empirical parameters to prevent the network from over-fitting.

In this work, we propose an online feedback method for data augmentation.
An illustration of the algorithm is shown in Fig. \ref{fig3b}. In the method,   the training data is  adjusted dynamically according to the actual effect in each iteration, which is beneficial to keep a balance on face scales. 
In our method, we construct a set $S$ which starts from all training samples for face detection. In each iteration, we  calculate the loss $L_{box}^s$ of face size less than  $25\times 25$.  The stitching strategy is used if the ratio of $L_{box}^s/L_{box}$ is less than value $T$.
 After each epoch, we test the current model on the training data and obtain all images with false positive and false negative boxes. Then we update $S$ with these difficult images.  In this way, the proposed method improves the balance of the data automatically. 

%
%
%
%

\section{Experiments} \label{experiments}
\subsection{Implementation details}
\textbf{Training Dataset:}The first dataset we use  is the WIDER FACE dataset \cite{Yang2016WIDER}.
It consists of 32,203 images and 393,703 face bounding boxes. As the original dataset does not contain pose labels, we first annotate the pose for the dataset. Head pose annotation from 2D images is an extremely challenging and time-consuming task, especially for very small faces. In our annotation, we annotate  in a semi-supervised way and skip faces smaller than   $35\times 35$ pixels.  We first pre-label  the data through FSAnet \cite{2019FSA} and Hopenet \cite{2018Fine}, and then two trained experts adjust  the pre-labelled annotations manually after careful examination and mutual agreement. In our annotation, we mainly adjust the Yaw, which is most important for DFR. To facilitate the research in this area, we are preparing the data for release and the data will be available at https://github.com/lyp-deeplearning/MOS-Multi-Task-Face-Detect. 

\textbf{Evaluation Metrics:}The AFLW \cite{koestinger2011annotated}  dataset is used to test the landmark localization accuracy. It contains 25993 faces with up to 21 landmarks per image. The evaluation is based on the mean errors measured by the distances between the estimated landmarks and the ground-truth, normalized with the face box size $\sqrt{w\times h}$.  

We conduct testing of Head Pose estimate on AFLW2000 \cite{zhu2016face} dataset. It  provides ground-truth 3D faces with pose angles for the first 2,000 images of the AFLW dataset. We follow the protocol of Hopenet ~\cite{2018Fine}.

\textbf{Model architecture and training protocols: } We implement  MOS-S based on lightweight backbone ShuffleNet V2 \cite{ma2018shufflenet}. 
MOS-S  employs three feature pyramid levels from $P3$ to $P5$. The backbones are pretrained on the Imagenet. We use online hard example mining  \cite{shrivastava2016training} and constrain the ratio of positive and negative anchors to 1:3. The  SSH \cite{najibi2017ssh} module is  added   to increase the receptive field.
The anchor aspect ratio is set to 1:1.  
In  MOS-S, we train the model with 640$\times$640 input images,  the feature pyramid is set
to $P3(80\times80)$, $P4(40\times40)$, $P5(20\times20)$, which leads to  16800 anchors.  
Although we   mainly focus  on the lightweight backbones, we also implement MOS-L with heavy  backbone ResNet152 \cite{he2016deep} to explore the performance of the proposed modules in this situation. 


We train the face detection networks with a batch size of 64 using 1 NVIDIA
Tesla V100 GPUs. We use SGD optimizer with momentum at 0.9, weight decay at 0.0005. The initial learning rate is set to 0.001 and then divided by 10 at 150 and 180 epochs. The training terminates at 200 epochs. For inference, we use multi-scale testing strategy as in \cite{zhang2017s3fd}.


\subsection{Ablation Study}
To justify the effectiveness of   each component in MOS-S, we conduct the following ablation studies. The baseline approach is based on ShuffleNet V2 and SSH without MTH, as shown in Fig. \ref{fig2} and Fig. \ref{fig3a}. Then we add the four modules including MTH, Pose loss, UML, Online Feedback and the commonly used deformable convolution network (DCN) \cite{dcnv2} one by one. The results are shown in Table \ref{table2}.
%
%

\begin{table}[] \footnotesize 
	\caption{Ablation study of the proposed modules on the WIDER FACE validation set and AFLW2000 ~\cite{zhu2016face}. The baseline adopts the the hard connect structure in Fig. \ref{fig3a} with MOS-S.}
	\begin{center}	
		\begin{tabular}{lcccc}\hline
			{Method}  & Easy          & Medium          & Hard   & Avg. MAE   \\      
			\hline
			Baseline              & 89.32        & 88.45        & 80.20   & 6.73              \\
			{+  MTH}              & 90.11        & 89.31         & 81.64  & 6.49       \\
			{+ UML}         & 91.18         & 90.26          & 83.01    & 6.14      \\
			{+ Online Feedback}   & 92.41         & 91.37          & 85.24   & 6.09       \\
			{+ DCN}          & 92.91      & 91.61      & 85.72   & 5.97      \\
			\hline	
		\end{tabular}
	\end{center}
	\label{table2}
	
\end{table}

\textbf{Multi-task head.} To justify the effectiveness of MTH, we first add MTH into the baseline, denoted as +MTH. As shown in Table \ref{table2},  it obtains 0.79\%, 0.86\%, and 1.44\%  improvement on the \textit{easy}, \textit{medium} and \textit{hard} subsets respectively for lightweight backbones.  

\textbf{Uncertainty Multi-task Loss.} The baseline uses heuristic weights ($L_{cls}:L_{box}:L_{pts}:L_{pose}=2:1:1:0.25$) for multi-tasking. Table. \ref{table2} compares the results of face detection and pose estimation tasks using UML. The UML loss further improves the performance by 1.07\%, 0.95\% and 1.37\%.  Simultaneously, this method also improves the result of head pose estimation from 6.49 to 6.14. 

\textbf{Online Feedback data-augmentation.} In the online feedback, we  add the samples with small losses and false positive. The results are denoted as +Online Feedback.   Table \ref{table2} show that it obtains an improvement of 1.23\%, 1.11\% and 2.23\% for \textit{easy}, \textit{medium} and \textit{hard} subsets respectively.  This shows that  the proposed feed-back strategy improves the balance  of the data and therefore the accuracy of face detection. In our experiments, we have observed that arbitrary increase of small faces would  bring the training  into a bottleneck as many false detection come from medium or large faces.




\begin{table}
	\footnotesize  
	\caption{Comparison of AP with other methods for light networks in the WIDER FACE validation and test set. * indicates the work without peer review. Blue indicates the best result with lightweight backbone.}
	\begin{center}
		\begin{tabular}{lccccc}\hline
			{Method}    & backbone & \textit{easy(val$\lVert$test)} & \textit{medium(val$\lVert$test)} & \textit{hard(val$\lVert$test)} &  {\textit{Multi-task}} \\  \hline
			MTCNN\cite{zhang2016joint}       & Customized  & 0.851 $\lVert$ $\ \ -\ \ \ \ $  & 0.820 $\lVert$ $\ \ -\ \ \ \ $  & 0.607 $\lVert$ $\ \ -\ \ \ \ $   & \Checkmark   \\
			{FACEBOXES}\cite{zhang2017faceboxes} & Customized  & 0.879 $\lVert$ 0.881 &  0.857 $\lVert$ 0.853     &  0.771 $\lVert$ 0.774 & \XSolid        \\
			{lffd v1*}\cite{he2019lffd}      & Customized   & 0.910 $\lVert$ 0.896  &  0.881 $\lVert$ 0.865  &  0.780 $\lVert$ 0.770  & \XSolid       \\
			{ASFD-D0*}\cite{zhang2020asfd}        & Customized   & 0.901 $\lVert$ $\ \ -\ \ \ \ $  &  0.875 $\lVert$ $\ \ -\ \ \ \ $  &  0.744 $\lVert$ $\ \ -\ \ \ \ $   & \XSolid       \\
			{RetinaFace}\cite{deng2019retinaface}         & MobileNet-0.25   & {0.914} $\lVert$ $\ \ -\ \ \ \ $  & {0.901} $\lVert$ $\ \ -\ \ \ \ $   &  {0.782} $\lVert$ $\ \ -\ \ \ \ $   & \Checkmark      \\
			{img2pose*}\cite{albiero2020img2pose}      & ResNet-18   & 0.908 $\lVert$ 0.900  &  0.899 $\lVert$ 0.891  &  0.847 $\lVert$ 0.839   & \Checkmark       \\
			MOS-S         & ShuffleNet V2         &\color{blue}{0.929 $\lVert$ 0.922}      &\color{blue}{0.916 $\lVert$ 0.911}        &\color{blue}{0.859 $\lVert$ 0.857}   & \Checkmark          \\ 
			MOS-L           & ResNet-152       &{0.969 $\lVert$ 0.955}        &{0.961 $\lVert$ 0.952}       &{0.921 $\lVert$ 0.913}    &\Checkmark        \\ 
			\hline
		\end{tabular}
	\end{center}	\label{table4}  
\end{table}
\subsection{Benchmark Results}
\textbf{Face Detection Accuracy. }We train the model on the training set and test on the WIDER FACE validation and test sets.  We follow the standard practices of \cite{najibi2017ssh, zhang2017s3fd} and employ
flip as well as multi-scale  strategies. The standard  Average Precision (AP) is computed. Table \ref{table4} shows the comparison   in the validation set. Taking the lightweight ShuffleNet V2 as backbone,   MOS-S    achieves AP of 92.9\%, 91.6\% and 85.9\%   on the three subsets respectively, surpassing other methods.    Taking the lightweight Mobilenet V2 ~\cite{sandler2018mobilenetv2} as backbone,   MOS-Mobilenetv2    achieves AP of 94.19\%, 93.25\% and 88.34\%   on the three subsets respectively which confirms that our method is also effective on other lightweight backbones. MOS-S has achieved trade-off between speed and accuracy.

Using the ResNet152 as backbone,  MOS-L achieves  AP of 96.9\%, 96.1\%, 92.1\% on the three subsets respectively.   

\begin{figure*} 
	\centering
	{\subfigure[Accuracy  of landmark detection]{ 
			\includegraphics[width=2.4in,height=1.0in]{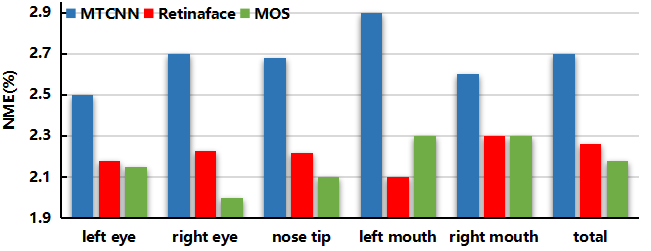}  \label{fig-zhuxinga}
	}}
	{\subfigure[Pose angle distribution in WIDER FACE training set]{ 
			\includegraphics[width=2.0in,height=1.0in]{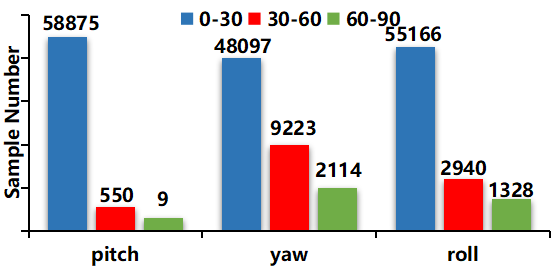}  \label{fig-zhuxingb}
	}}	
	\caption{(a) The comparison of MOS-S with other multi-task face detector on AFLW dataset.  (b) The pose angle distribution in WIDER FACE training set. We can observe that majority of face angles are concentrated below 30 degrees.  }
	\label{fig-zhuxing}
\end{figure*}

\textbf{Landmark Localization Accuracy. }To evaluate the accuracy of the landmark localization, we compare MOS-S with the commonly used MTCNN \cite{zhang2016joint} and RetinaFace \cite{deng2019retinaface} with landmark output.  We train MOS-S on the WIDER FACE and test on the AFLW \cite{koestinger2011annotated} dataset.  Fig. \ref{fig-zhuxinga} compares the proposed MOS-S with RetinaFace and MTCNN. As we can see, MOS-S performs the best.

%

\begin{minipage}{\textwidth}\scriptsize
	\vspace{0.3cm}
	
	\begin{minipage}[t]{0.48\textwidth}
		\makeatletter\def\@captype{table}
		\caption{Evaluation on AFLW2000. }
		\begin{tabular}{lcccccc}
			\hline
			{Method} & \begin{tabular}[c]{@{}c@{}}Yaw \end{tabular} & \begin{tabular}[c]{@{}c@{}}Pitch \end{tabular} & \begin{tabular}[c]{@{}c@{}}Roll \end{tabular}  & MAE\\ 	\hline
			{Dlib(68 points) \cite{kazemi2014one}}  & 23.1       &13.6                                                &10.5                                               & 15.8   \\
			{Fan(12 points) \cite{bulat2017far}}  &6.36                                               &12.3                                                 & 8.71                                                & 9.12   \\
			{3DDFA \cite{zhu2016face}}  & 5.40                                               &8.53                                                 &8.25                                               & 7.39  \\ 
			{Hopenet \cite{2018Fine} ($\alpha=2$)}  & 6.47                                                & 6.56                                                & 5.44                                                & 6.16                                                  \\
			{FsaNet \cite{2019FSA} }  & 4.50                                                & 6.08                                                 & 4.64                                                & 5.07                                                  \\
			{QuatNet \cite{hsu2018quatnet} }  & \color{blue}3.97                                                & \color{blue}{5.62}                                                 & \color{red}{3.92}                                                & \color{blue}4.50                                                  \\
			\textbf{MOS-S-widerface}  
			& 4.52                                                & 6.91                                                 & 6.48                                                & 5.97                                                    \\
			\textbf{MOS-L-widerface}  
			& 4.05                                                & 6.29                                                 & 5.87                                                & 5.40                                                    \\
			\textbf{MOS-S-300wlp}  & \color{red}{3.91}                                                & \color{red}5.42                                                 & \color{blue}3.98                                                & \color{red}4.43                                                    \\
			\hline
		\end{tabular}
		
		\label{table5}
	\end{minipage}
	\begin{minipage}[t]{0.48\textwidth}
		
		\makeatletter\def\@captype{table}
		\caption{Avearge inference time  in WIDER FACE validation set.}
		\begin{tabular}{lcc}
				\hline
			Method    & Time (ms)    \\ \hline
			
			{RetinaFace-M\cite{deng2019retinaface}}+ PnP \cite{pnp1981}(GPU)    & 7.2    \\
			{RetinaFace-M\cite{deng2019retinaface}} + Hopenet \cite{2018Fine}(GPU)    & 47.9   \\
			{RetinaFace-M\cite{deng2019retinaface}} + FSAnet \cite{2019FSA}(GPU)    & 41.1   \\
			
			\textbf{MOS-S}(GPU)    & 10.8   \\
			\textbf{MOS-S}(ARM)    & 68   \\
			\textbf{MOS-L}(GPU)    & 62.2    \\
			\hline
		\end{tabular}
		
		\label{table6}
	\end{minipage}
\vspace{0.3cm}
\end{minipage}


\textbf{Head Pose Estimation Accuracy. }We evaluate the   pose estimation  using the AFLW2000 \cite{zhu2016face}. Of the six methods \cite{kazemi2014one,bulat2017far,zhu2016face,2018Fine,2019FSA,hsu2018quatnet} compared, MOS-S is the only method to predict face location and head pose simultaneously. 
From   Table \ref{table5}, we observe that MOS-S-widerface  performs good in Yaw, but worse in Pitch and Roll. The lower performance of in Pitch and Roll estimation is likely because of few samples with large pitch and roll angles in the   training data. Fig. \ref{fig-zhuxingb} plots the distribution of the angles. As we can see, the WIDER FACE dataset has few faces with large pitch and roll angle. 

Note that the comparison here is not a completely fair one as MOS is trained using WIDER FACE while others are trained using 300W-LP \cite{zhu2016face} which include various distribution angle data. For fair comparison, we add another  experiment by training  the model using  only   the 300W-LP dataset. MOS-S-300wlp in Table \ref{table5} achieves state-of-the-art accuracy and minimal inference time. 

\begin{figure*} 
	\centering
	{\subfigure{ 
			\includegraphics[width=1.4in,height=0.76in]{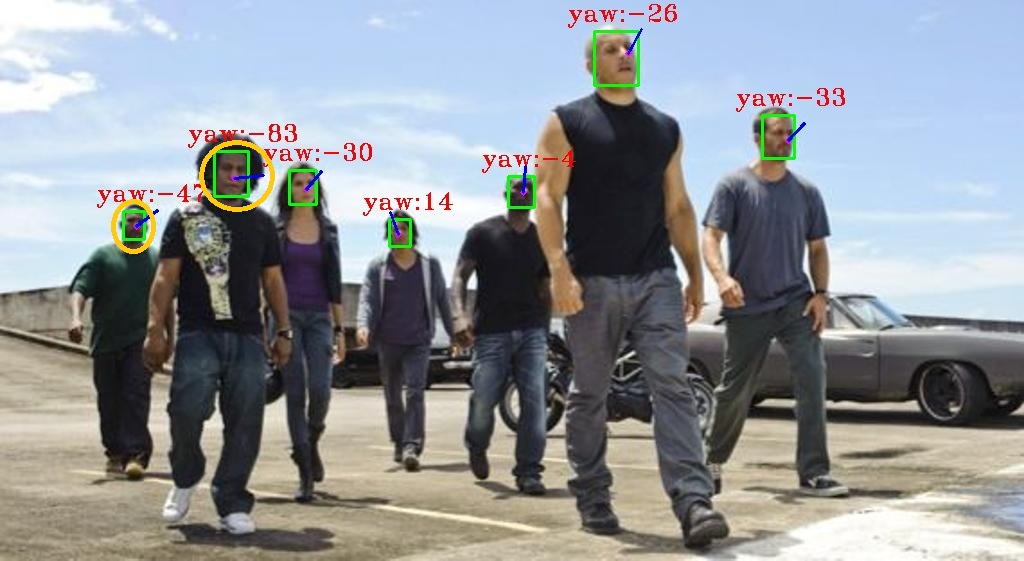}
	}} 
	{\subfigure{ 
			\includegraphics[width=1.4in,height=0.76in]{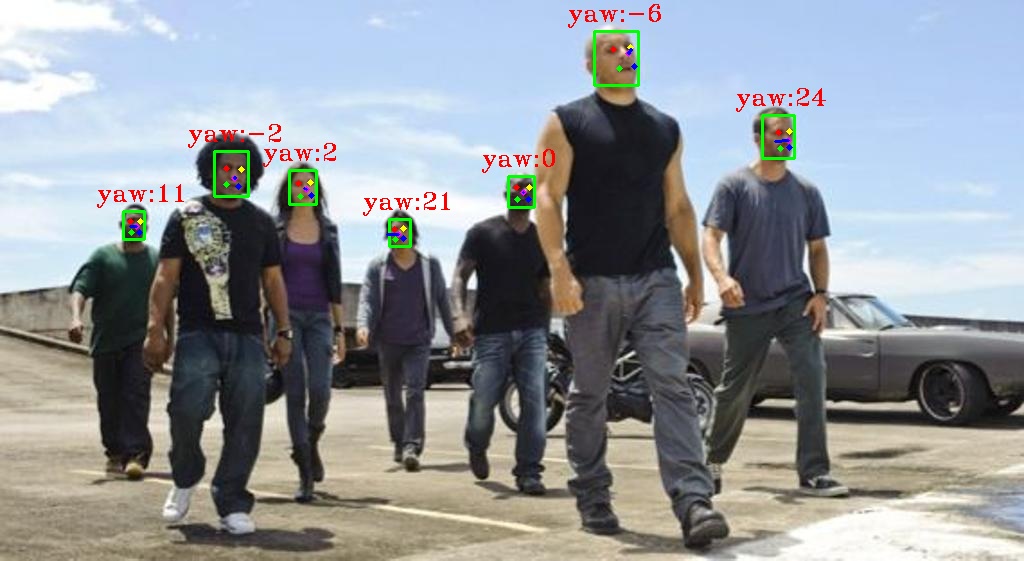}
	}}  
	{\subfigure{ 
			\includegraphics[width=1.4in,height=0.76in]{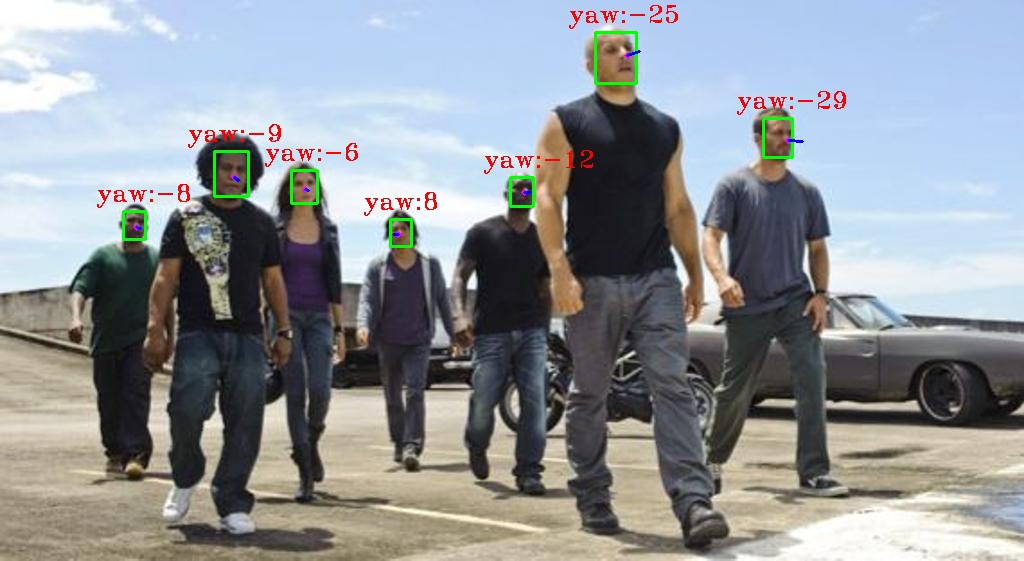}
	}} \vspace{-0.1in}  \\ \setcounter{subfigure}{0}
	{\subfigure[RetinaFace+PnP \cite{pnp1981}]{ 
			\includegraphics[width=1.4in,height=0.76in]{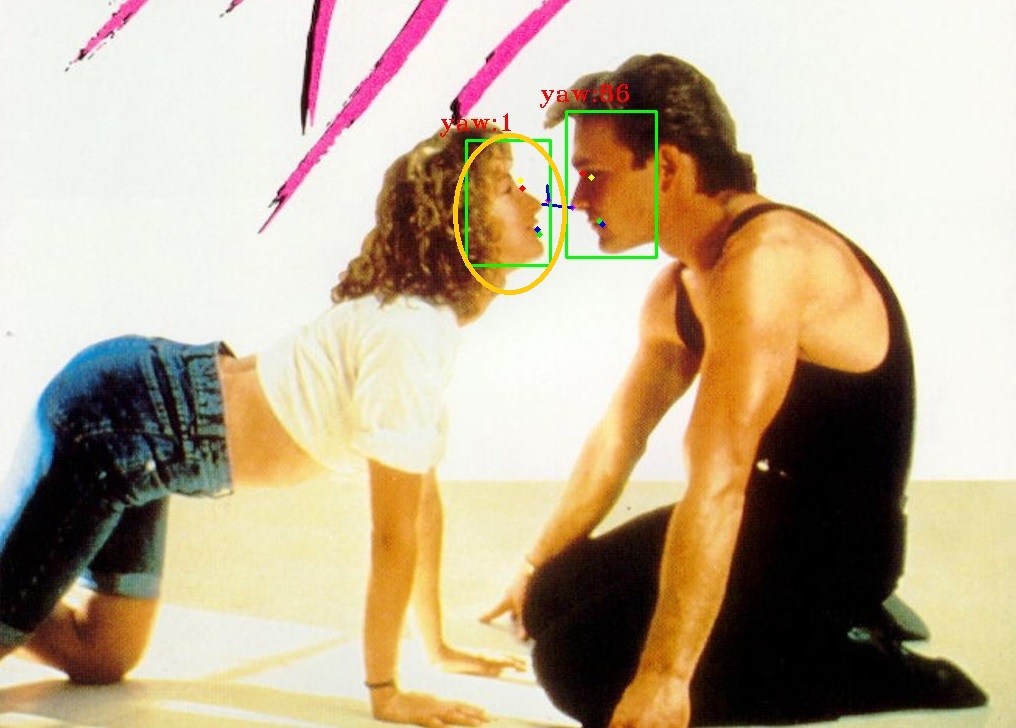}
	}}
	{\subfigure[RetinaFace+Hopenet \cite{2018Fine}]{ 
			\includegraphics[width=1.4in,height=0.76in]{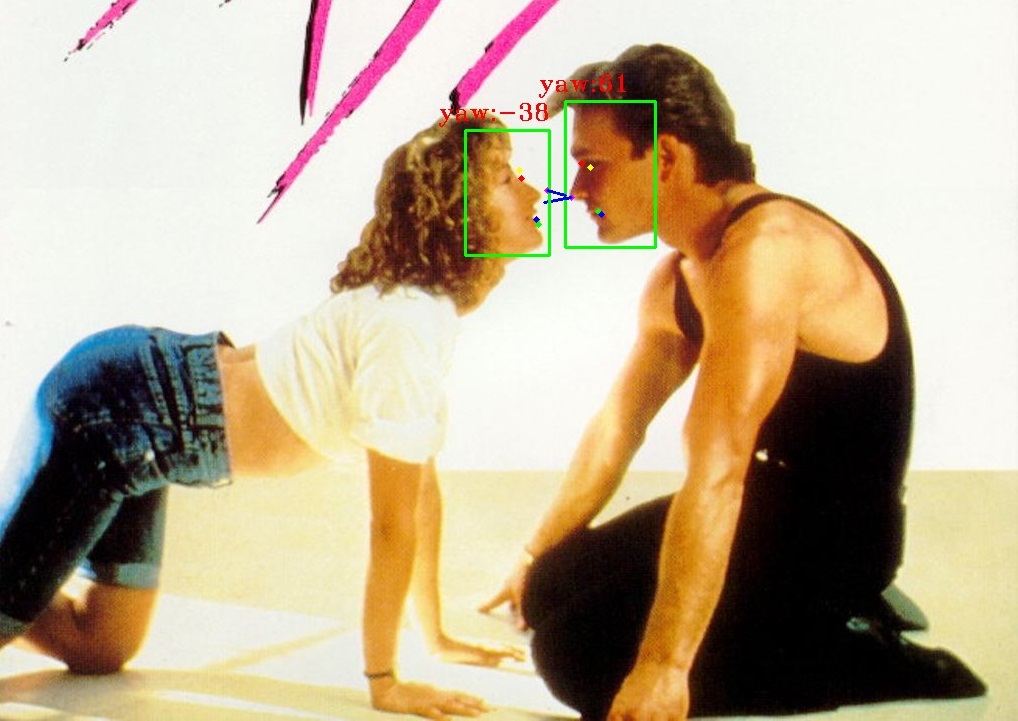}
	}}
	{\subfigure[MOS-S]{ 
			\includegraphics[width=1.4in,height=0.76in]{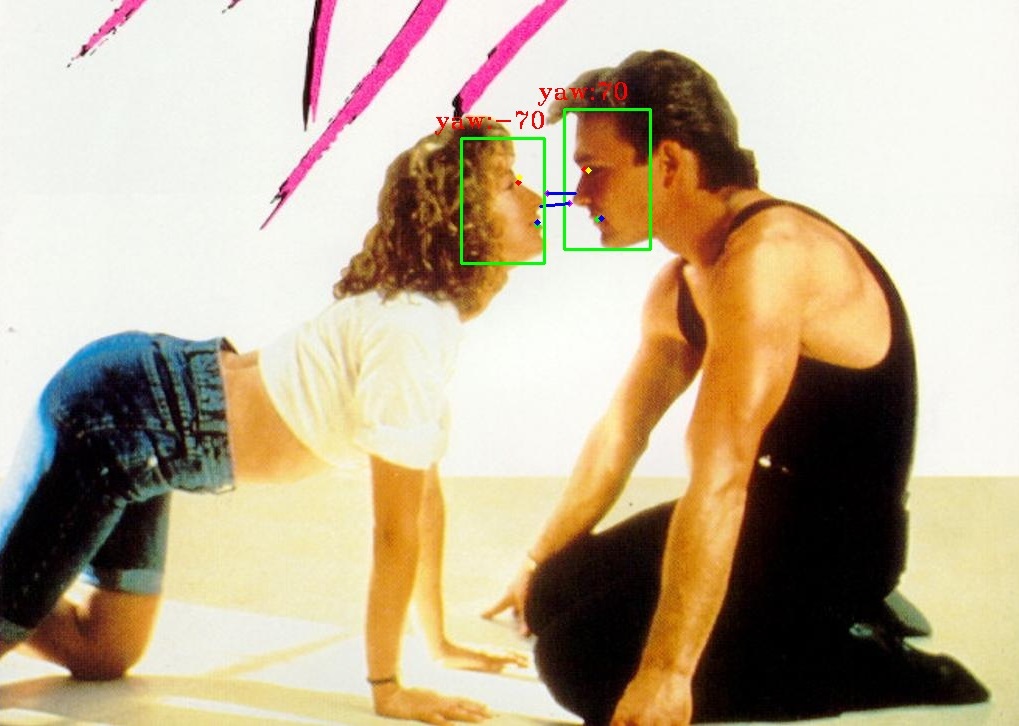}
	}}	
	\caption{Visual comparison of results between MOS-S and RetinaFace (MobileNet-0.25 backbone) with different pose methods. We only show the yaw angle which is most important in DFR, and the yellow circle indicates cases with large errors. It takes 7.6ms, 49.0ms, and 12.1ms to process the images in the first row by the three different methods respectively.  }
	\label{fig9p}
\end{figure*}

Some examples have been given to compare MOS-S with RetinaFace with different pose estimation methods in Fig. \ref{fig9p}. As we can see, PnP \cite{pnp1981} performs poorly while Hopenet \cite{2018Fine} requires more computational time. Our method is able to provide accurate results efficiently.

\textbf{Comparison with Hyperface. }For a fair comparison, we use AFLW as the training set, same as that in Hyperface \cite{ranjan2017hyperface}. We compared MOS-S and Hyperface-alexnet in face detection and pose estimation. For face detection,  MOS-S achieves mAP of 93.2\% on the FDDB dataset\cite{jain2010fddb}, which is better than 90.1\% by Hyperface-alexnet. For pose estimation task, MOS-S achieves 4.89 MAE better than 5.88 by Hyperface-alexnet in the AFLW dataset.


\subsection{Inference Efficiency}

In order to better compare the results in dynamic face detection, MOS-S is compared with RetinaFace-M( MobilenetV1\cite{howard2017mobilenets}) combined with Hopenet or FSAnet. The average time for images (resized to $640 \times 480$) from the WIDER FACE validation set are summarized in Table \ref{table6}. We measure inference time of MOS  using RTX 2060 GPU and ARM platform(RK3399).   With lightweight backbones,  MOS-S outperforms state of the art in face detection and pose estimation with less inference time.   
With heavy backbones, MOS-L  takes about 62.2 ms. 
Besides the above, we have also deployed  MOS-S using ncnn with multi-threads  on ARM platform and it achieves 15 FPS for a mobile device.


\section{Conclusion}
Real-time face detection, landmark localization and head pose estimation with low computational resource are challenging tasks. In this work, we propose a novel low latency and lightweight backbone to learn   the three tasks simultaneously, which  facilitates the computation for DFR in mobile devices such as robots. Uncertain multi-task loss has been proposed to regularize the learning. Moreover, we propose an online feedback sampling    to augment the data according to the performance of the trained models in the training iterations.   The experimental results show that the proposed method achieves the state-of-the-art results compared with other methods with similar computation resources.  Our codes and annotation will be made publicly available to facilitate further research in the area.

\textbf{Acknowledgment:} The work is supported in part by  Key-Area Research and Development Program of Guangdong Province, China, under grant 2019B010154003, and the Program of Guangdong Provincial Key Laboratory of Robot Localization and Navigation Technology, under grant 2020B121202011.

\bibliography{egbib}
\end{document}